\documentclass[10pt,twocolumn,letterpaper]{article}

\usepackage{cvpr}
\usepackage{times}
\usepackage{epsfig}
\usepackage{graphicx}
\usepackage{amsmath}
\usepackage{amssymb}
\usepackage[T1]{fontenc}

\usepackage[sort,numbers]{natbib}

\usepackage{xparse}
\NewDocumentCommand{\row}{>{\SplitArgument{2}{,}}m}{%
  \finalR#1%
}
\NewDocumentCommand{\finalR}{mmm}{%
  #1 & #2 & #3\\%
}
\NewDocumentCommand{\osize}{>{\SplitArgument{3}{,}}m}{%
  \finalOsize#1%
}
\NewDocumentCommand{\finalOsize}{mmmm}{%
  #1$\times$#2$\times$#3$\times$#4%
}

\usepackage{multirow,enumitem,booktabs}
\usepackage[table]{xcolor}
\usepackage{algorithm2e}
\usepackage{bbm}
\makeatletter
\newcommand{\removelatexerror}{\let\@latex@error\@gobble}
\makeatother

\newcommand{\ourso}{Ours}

\usepackage[pagebackref=true,breaklinks=true,letterpaper=true,colorlinks,bookmarks=false]{hyperref}

\cvprfinalcopy

\begin{document}

% Just placeholder for now
\title{Generating and Exploiting Probabilistic Monocular Depth Estimates}

\author{Zhihao Xia$^1$, Patrick Sullivan$^2$, Ayan Chakrabarti$^1$\\
  $^1$Washington University in St. Louis~~~~~$^2$The Boeing Company\\
  {\tt\small zhihao.xia@wustl.edu, patrick.l.sullivan2@boeing.com, ayan@wustl.edu}
}

\maketitle

\begin{abstract}
  Beyond depth estimation from a single image, the monocular cue is useful in a broader range of depth inference applications and settings---such as when one can leverage other available depth cues for improved accuracy. Currently, different applications, with different inference tasks and combinations of depth cues, are solved via different specialized networks---trained separately for each application. Instead, we propose a versatile task-agnostic monocular model that outputs a probability distribution over scene depth given an input color image, as a sample approximation of outputs from a patch-wise conditional VAE. We show that this distributional output can be used to enable a variety of inference tasks in different settings, without needing to retrain for each application. Across a diverse set of applications (depth completion, user guided estimation, etc.), our common model yields results with high accuracy---comparable to or surpassing that of state-of-the-art methods dependent on application-specific networks.
\end{abstract}

\section{Introduction}
\label{sec:intro}

Monocular depth estimation methods---that predict scene depth from only a single color image---have achieved surprising success through the use of deep neural networks~\cite{eigen2015predicting, wang2015towards, chakrabarti2016depth, fu2018deep, laina2016deeper}. This success confirms that even a single view contains considerable information about scene geometry. Purely monocular depth map estimates, however, are far from being precisely accurate given the ill-posed nature of the task. Fortunately, many practical systems are able to rely on other (yet also imperfect) sources of depth information---limited measurements from depth sensors, interactive user guidance, consistency across frames or views, etc. And so, it is desirable to combine these other sources with the monocular cue to extract depth estimates that are more accurate than possible from one source alone.

\begin{figure*}[!t]
  \begin{center}
    \includegraphics[height=13.1em]{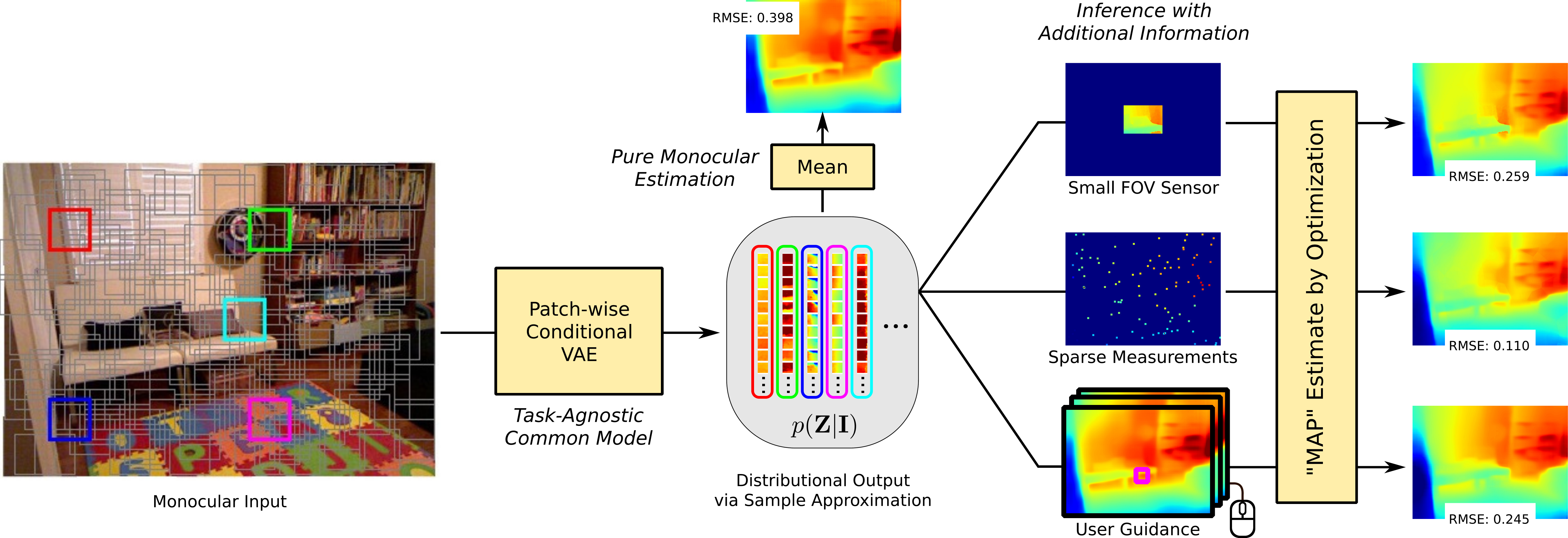}
  \end{center}
  \caption{Overview of our approach. Given an input color image, we use a common task-agnostic network to output a joint probability distribution $p(\mathbf{Z}|\mathbf{I})$ over the depth map---formed as a sample approximation using outputs of a conditional VAE that generates plausible estimates for depth in overlapping patches. The mean of this distribution represents a standard monocular depth estimate, but the distribution itself can be used to solve a variety of inference tasks in different application settings---including leveraging additional depth cues to yield improved estimates. All these applications are enabled by a common model, \emph{that is trained only once}.}
  \label{fig:teaser}
\end{figure*}

Although the monocular \emph{cue} is useful for augmenting other depth cues, the same isn't true for monocular \emph{estimators} that simply output a depth map, a form which can not be directly combined with additional depth cues. Instead, researchers have treated depth estimation using different combinations of cues as different applications in their own right (\eg, depth up-sampling~\cite{chen2018estimating}, estimation from sparse~\cite{ma2018sparse} and line~\cite{liao2017sparse} measurements, etc.), and solved each by learning separate estimators that take their corresponding set of cues, in addition to the color image, as input. This requires, for each application, determining the types of inputs that will be available, constructing a corresponding training set, choosing an appropriate network architecture, and then training that application-specific network---a process that is redundant and often onerous.

In this paper, we introduce a universal and versatile network to leverage the monocular depth cue in multiple application settings \emph{without re-training}. Our network is trained in an application-agnostic way on image-depth pairs, but can be utilized for inference in different applications and combined with different external depth cues as illustrated in Fig.~\ref{fig:teaser}. Rather than producing a depth map estimate, our monocular network outputs a \emph{probability distribution} over scene depth given an input color image. This distribution faithfully encodes both the information and ambiguity of depth values and their spatial dependencies based on the monocular input, and is produced in a form that can be combined with other depth cues during inference.

Our contributions are as follows:
\begin{itemize}[nosep,leftmargin=\labelwidth]
\item We propose a novel approach to output a probability density function that can express arbitrary beliefs and spatial dependencies for depth, conditioned on the image input. We train a conditional VAE~\cite{condVAE} to output multiple plausible depth samples independently for individual overlapping patches, and form the density as a sample approximation from all samples and patches.
\item We describe an efficient optimization method for inference that combines this image-conditional density function with other sources of depth information (e.g., from sensors or user input).
\item We show that our probabilistic outputs are useful for general inference tasks beyond depth map estimation---e.g., predicting pairwise ordinal depth relationships.
\item We carry out extensive experiments on the NYUv2 dataset~\cite{silberman2012indoor} to demonstrate the efficacy of our approach on a diverse variety of applications. All applications are enabled by the same network that is trained only once, but delivers accuracy comparable to or surpassing state-of-the-art methods dependent on task-specific models.
\end{itemize}

\section{Related Work}
\label{sec:rw}

\noindent\textbf{Monocular Depth Estimation.} First attempted by Saxena \etal~\cite{saxena2006learning}, early work in estimating scene depth from a single color image relied on hand-crafted features~\cite{saxena2009make3d,ladicky2014pulling,shi2015break,ranftl2016dense}, use of graphical models~\cite{saxena2009make3d,liu2014discrete,zhuo2015indoor}, and databases of exemplars~\cite{konrad2013learning,karsch2014depth}. More recently, 
Eigen \etal~\cite{eigen2014depth} showed that, given a large enough database of image-depth pairs~\cite{silberman2012indoor}, convolutional neural networks could be trained to achieve significantly more reliable depth estimates. Since then, there have been steady gains in accuracy through the development of improved neural network-based methods~\cite{eigen2015predicting,zhang2015monocular, wang2015designing,roy2016monocular,liu2016learning,chakrabarti2016depth,Li_2017_ICCV,heo2018monocular,lee2018single,fu2018deep}, as well as strategies for unsupervised an semi-supervised learning~\cite{garg2016unsupervised,kuznietsov2017semi,chen2016single}. Beyond estimating absolute depth, some works have also looked at pairwise ordinal depth relations between pair of points in the scene from a input color image~\cite{zoran2015learning,chen2016single}.

\noindent\textbf{Probabilistic Outputs.} Monocular depth estimators commonly output a single estimate of the depth value at each pixel, hindering their use in different estimation settings. Some existing methods do produce distributional outputs, but as per-pixel variance maps~\cite{kendall2017uncertainties,heo2018monocular} or per-pixel probability distributions~\cite{liu2019neural}. Note that depth values at different locations are not statistically independent, i.e., different values at different locations may be plausible independently, but not in combination. Thus, per-pixel distributions provide only a limited characterization that, while useful in some applications, can not be used more generally, \eg, to spatially propagate information from sparse measurements. 

Beyond per-pixel distributions, Chakrabarti \etal~\cite{chakrabarti2016depth} train a network to produce independent distributions for different local depth derivatives. They describe a method to use these derivative distributions to generate a better estimate of global depth, but do not provide a way to solve other tasks. Also, since their network output is restricted to uni-variate distributions for hand-chosen derivatives, it can not express the general spatial dependencies in a joint distribution over depth that we seek to encode for inference.

\noindent\textbf{Depth from Partial Measurement.} Since making dense depth measurements is slow and expensive, it is useful to be able to recover a high-quality dense depth map from a small number of direct measurements by exploiting the monocular cues in a color image. A popular way of combining color information with partial measurements is by requiring color and depth edges to co-occur: this approach is often successful for ``depth inpainting'', i.e., filling in gaps of missing measurements in a depth map (common in measurements from structured light sensors). A notable and commonly-used example is the colorization method of Levin \etal~\cite{levin2004colorization}. Other methods along this line include \cite{herrera2013depth,liu2012guided,liu2013guided,Matsuo_2015_CVPR,doria2012filling}, while Zhang and Funkhouser~\cite{zhang2018deep} used a neural network to predict normals and occlusion boundaries to aid inpainting.

However, when working with a very small number of measurements, the task is significantly more challenging (see discussion in \cite{chen2018estimating}) and requires relying more heavily on the monocular cue. In this regime, the solution has been to train a network that takes the color image and the provided sparse samples as input. Various works have adopted this approach for measurements along a single horizontal line from a line sensor~\cite{liao2017sparse}, random sparse measurements~\cite{van2019sparse,ma2018sparse,jaritz2018sparse,shivakumar2019dfusenet}, and sub-sampled measurements on a regular grid~\cite{li2016deep,gu2017learning,chen2018estimating}. Note that several of these methods also train separate networks even for different settings of the same application, such as for different sparsity levels~\cite{ma2018sparse} and different resolution grids~\cite{chen2018estimating}.

An exception here is the depth completion method of Wang \etal~\cite{wang2019plug} who use a pre-trained monocular depth network, and provide a way to improve its monocular predictions when given sparse depth measurements. They iteratively back-propagate errors between measurements and the network output to update activations of an intermediate layer (but not the network weights), leading to an improved depth map output. Thus, their method uses the monocular network's output as an initialization, and its internal representation as a structured way to spatially propagate measurement information. In contrast, our method outputs an explicit probabilistic representation which can be used for depth completion as well as for other inference tasks, and as our experiments show, yields more accurate results.

\noindent\textbf{Networks for Generating Samples.} In this work, we form a conditional joint distribution of depth values by training our network to generate samples of multiple plausible depth values. In particular, we follow the approach of \cite{condVAE} to train a conditional VAE and use its outputs to form a sample approximation to the joint distribution. Note that instead of generating samples of a global map (like in \cite{condVAE}), we train the VAE to produce samples for individual overlapping patches independently. We also conduct ablation experiments using a conditional GAN~\cite{goodfellow2014generative,mirza2014conditional} to produce these samples, and while the VAE formulation performs better, our results with the GAN are also reasonable. This suggests our approach is able to exploit any neural network-based method for generating conditional samples, and can benefit from future advances in this direction.

\newcommand{\paras}[1]{\noindent\textbf{#1}}
\renewcommand{\Phi}{\mathcal{S}}
\newcommand{\crop}{\mathcal{P}_i}
\newcommand{\patch}{\mathbf{x}}
\newcommand{\Z}{\mathbf{Z}}
\newcommand{\I}{\mathbf{I}}
\newcommand{\A}{\mathbf{A}}

\section{Proposed Method}
\label{sec:net}

Given the RGB image $\I$ of a scene, our goal is to reason about its corresponding depth map $\Z\in \mathbb{R}^N$, represented as a vector containing depth values for all $N$ pixels in the image. Rather than predict a single estimate for $\Z$, we seek to output a \emph{distribution} $p(\Z|\I)$, to more generally characterize depth information and ambiguity present in the image. In this section, we describe our approach for generating this distributional output, and equally importantly, for exploiting it for inference in various applications.

\subsection{Probabilistic Monocular Depth}

We form the distribution $p(\Z|\I)$ as a product of functions defined on individual overlapping patches as
\begin{equation}
  \label{eq:factor}
  p(\Z|\I) \propto \prod_i \psi_i(\crop{}\Z | \I),
\end{equation}
where $\psi_i(\cdot)$ is a potential function for the $i^{th}$ patch, and $\crop{}$ a sparse matrix that crops out that patch from $\Z$ (for patches of size $K\times K$, each $\crop{}$ is a $K^2\times N$ matrix). Note that this is a Markov Random Field with $K\times K$ patches as the maximal cliques, and since these patches overlap, depth values at all pixels---including those that do not lie in the same patch---are statistically inter-dependent.

\paras{Generating Samples.} To form the per-patch potentials $\psi_i(\cdot)$, we train a network that produces samples of depth given the image input, and run it multiple times during inference to generate multiple plausible samples. A crucial aspect of this network is that, instead of  sampling the global depth map, it generates separate samples independently for the depth $\crop{}\Z$ of every patch $i$. This ensures that depth values within each sample represent a plausible estimate for the corresponding patch, but that samples of different patches are conditionally independent given the image. Limiting the dimensionality of each sample allows us to approximate the per-patch potential $\psi_i(\cdot)$ with a reasonable number of samples, while enforcing independence between samples of different patches ensures that the overall distribution $p(\Z|\I)$ in \eqref{eq:factor} sufficiently captures the global ambiguity in depth.% $\Z$.
\begin{figure}[!t]
  \begin{center}
    \includegraphics[width=0.9\columnwidth]{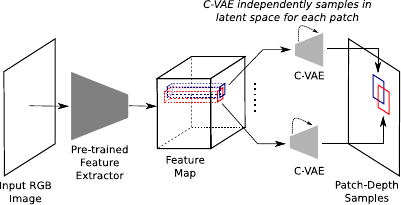}
  \end{center}
  \caption{Generating samples with a conditional VAE. Our network generates samples for depth independently in each overlapping patch, and we run it multiple times to generate multiple plausible samples per-patch. The input to the VAE comes from pre-trained feature extraction layers from a state-of-the-art monocular model~\cite{fu2018deep}. Samples generated for different patches (including those that overlap) are kept statistically independent---after conditioning on the image---by using separate per-patch latent vectors.}
  \label{fig:arch}
\end{figure}

We adopt the conditional VAE framework proposed in \cite{condVAE} for generating samples---that features a ``prior-net'' to predict distribution over values of a latent vector from the image, with an encoder-decoder network that predicts depth values from the image and a sample from this latent distribution. To reduce complexity, we bootstrap our network by taking a pre-trained state-of-the-art monocular depth estimation network (DORN~\cite{fu2018deep}), removing the last two convolution layers, and treating the remaining layers as a ``feature extractor''. These features, rather than the image itself, are provided as input to the conditional VAE.

We achieve patch independent sampling by having a separate latent vector for each patch. We set up the architecture of the decoder in the encoder-decoder network to produces an estimate of the depth of each overlapping patch using only its own latent vector,  and not those of overlapping patches. The prior-net is also setup to predict separate distributions for the latent vector of each patch (as is the posterior-net during training). At test time, we draw multiple samples independently from the latent space for each patch, which the encoder-decoder network uses to generate correspondingly independent per-patch depth samples. A more detailed description of the VAE architecture and training approach is included in the supplementary. 

\paras{Sample Approximation.} Next, given a set $\Phi_i$ of samples $\{\patch{}_i^s\}$ for each patch $i$, we define its potential $\psi_i(\cdot)$ as
\begin{equation}
  \label{eq:potdef}
  \psi_i(\crop{}\Z|\I) = \frac{1}{|\Phi_i|} \sum_{\patch{}_i \in \Phi_i} \exp\left(-\frac{\|\crop{}\Z - \patch{}_i\|^2}{2h^2}\right).
\end{equation}
This can be interpreted as forming a kernel density estimate from the depth samples in $\Phi_i$ using a Gaussian kernel, were the Gaussian bandwidth $h$ is a scalar hyper-parameter\footnote{While $h$ can be estimated based on the variance between $\patch{}_i$ and true patch depths, as we will see, its actual value is often not needed as it is factored into other manually-set, task-specific parameters.}.

Unlike independent per-pixel~\cite{kendall2017uncertainties,heo2018monocular,liu2019neural} or per-derivative~\cite{chakrabarti2016depth} distributions, the samples $\{\Phi_i\}$ enable the patch potentials $\psi_i(\cdot)$ to express complex spatial dependencies between depth values in local regions. Moreover, our joint distribution $p(\Z|\I)$ is defined in terms of overlapping patches, and thus models dependencies across the entire depth map. During inference, this enables information propagation across the entire scene, and reasoning about the global plausibility of scene depth estimates. 

Note that the distribution $p(\Z|\I)$ can be used to recover a monocular depth map estimate as the mean over $p(\Z|\I)$ by computing the average estimate of depth at each pixel from all samples from all patches that include that pixel. But the real utility of our distributional output comes from enabling a variety of inference tasks, as we describe next.

\subsection{Depth Estimation with Additional Information}
\label{sec:addinf}

In several applications, a system has access to additional sources beyond the monocular image that provide some partial information about depth. Our distributional output allows us to combine the monocular cue with these sources, and derive a more accurate scene depth estimate  than possible from either source alone. Specifically, we assume the additional depth information is provided in the form of a cost $C(\Z)$, and combine it with our distribution $p(\Z|\I)$ to derive a depth estimate $\hat{\Z}$ as:
\begin{eqnarray}
  \label{eq:map}
  \hat{\Z}=\arg \min_{\Z} -\log p(\Z|\I) + C(\Z),\qquad\qquad\notag\\
  \log p(\Z|\I)=\sum_i\log\sum_{\patch{}_i\in\Phi_i}\exp\left(-\frac{\|\crop{}\Z - \patch{}_i\|^2}{2h^2}\right).
\end{eqnarray}
With some abuse of terminology, this can be thought of as computing the maximum a posteriori (MAP) estimate of $\Z$, where $p(\Z|\I)$ is the image-conditional ``prior'', and $C(\Z)$ can be interpreted as a ``likelihood'' from the additional depth information source.

The log-likelihood of our distribution in \eqref{eq:map} can be simplified with a standard approximation of replacing the summation over exponentials with a maximum (since $\crop{}\Z$ is high-dimensional, the largest term typically dominates):
\begin{eqnarray}
  \label{eq:mapsimp}
  \hat{\Z}\approx \arg \min_{\Z} - \sum_i~\log~\max_{\patch{}_i\in\Phi_i}~\exp\left(- \frac{\|\crop{}\Z - \patch{}_i\|^2}{2h^2}\right)\hspace{-3em}~\notag\\+ ~C(\Z)\qquad\qquad \notag\\
   =\arg \min_{\Z}~\min_{\{\patch_i \in \Phi_i\}}~\sum_i \| \crop{}\Z - \patch{}_i\|^2 + 2h^2~C(\Z).
\end{eqnarray}
Note that this expression now involves a minimization over both $\Z$ and selections of samples $\patch{}_i \in \Phi_i$ for every patch.

We will use two forms of the external cost $C(\Z)$ to encode available information in various applications. The first is simply a generic global cost that we denote by $C^G(\Z)$, and the other is one that can be expressed as a summation over the depth values of individual patches $\sum_i C_i(\crop{}\Z)$. Including both these possible forms in \eqref{eq:mapsimp}, we arrive at the following optimization task:
\begin{equation}
  \label{eq:genopt}
  \min_\Z \min_{\{\patch{}_i\in\Phi_i\}} \sum_i \|\crop{}\Z - \patch{}_i\|^2+ \underbrace{\sum_i C_i(\patch{}_i) + C^G(\Z)}_{\text{Possible forms of } C(\Z)},
\end{equation}
where the factor $2h^2$ is absorbed in the definitions of the costs, and the per-patch costs $C_i(\crop{}\Z)$ are approximated as $C_i(\patch{}_i)$ to act on samples instead of crops of $\Z$ (we assume this will roughly be equivalent at convergence).

We use a simple iterative algorithm to carry out this optimization. The global depth $\Z$ is initialized to the mean per-pixel depth from $p(\Z|\I)$, and the following updates are applied alternatingly to $\{\patch{}_i\}$ and $\Z$ till convergence:
\begin{eqnarray}
  \patch{}_i &\leftarrow& \arg \min_{\patch{}_i \in \Phi_i} \|\crop{}\Z - \patch{}_i\|^2 + C_i(\patch{}_i),~~~\forall i.\label{eq:xupd}\\
  \Z &\leftarrow& \arg \min_{\Z} \|\crop{}\Z - \patch{}_i\|^2 + C^G(\Z).\label{eq:zupd}
\end{eqnarray}

The updates to patch estimates $\patch{}_i$ can be done independently, and in parallel, for different patches. The cost in \eqref{eq:xupd} is the sum of the squared distance from corresponding crop  $\crop{}\Z$ of the current global estimate, and the per-patch cost $C_i(\cdot)$ when available. We can compute these costs for all samples in $\Phi_i$, and select the one with the lowest cost. Note that the cost $C_i(\cdot)$ on all samples need only be computed once at the start of optimization.

The update to the global map $\Z$ in \eqref{eq:zupd} depends on the form of the global cost $C^G(\cdot)$. If no such cost is present, $\Z$ is given by simply the overlap-average of the currently selected samples $\patch{}_i$ for each patch. For applications that do feature a global cost, we find it sufficient to solve \eqref{eq:zupd} by first initializing $\Z$ to the overlap-average, and then carrying out a small number of gradient descent steps as
\begin{equation}
  \label{eq:gradzupd}
  \Z \leftarrow \Z - \gamma \nabla_{\Z} C^G(\Z),
\end{equation}
where the scalar step-size $\gamma$ is a hyper-parameter.

We now discuss concrete examples of our inference approach by considering specific applications, and describe associated choices of the costs $C^G(\cdot)$ and $C_i(\cdot)$.

\subsubsection{Depth Completion}
\paras{Dense Depth from Sparse Measurements.} We consider the task of estimating the depth map $\Z$ when an input sparse set $\mathbf{F}$ of depth measurements  at isolated points in the scene is available, along with a color image. We use the measurements $\mathbf{F}$ to define a global cost $C^G(\cdot)$ in \eqref{eq:genopt} as
\begin{equation}
  \label{eq:cgmeasure}
  C^G(\Z) = \lambda \|\Z\downarrow - \mathbf{F}\|^2,
\end{equation}
where $\downarrow$ represents sampling $\Z$ at the measured locations. Based on this, we define the gradients to be applied in \eqref{eq:gradzupd} for computing the global depth updates as
\begin{equation}
  \label{eq:gzupd}
  \nabla_{\Z}C^G(\Z) = \lambda (\Z\downarrow - \mathbf{F})\uparrow,
\end{equation}
where $\uparrow$ represents the transpose of the sampling operation. Since both the weight $\lambda$ and the step-size $\gamma$ in \eqref{eq:gradzupd} are hyper-parameters, we simply set $\lambda=1$, and set the step-size $\gamma$ (as well as number of gradient steps) based on a validation set.

We consider two kinds of sparse inputs. The first are at arbitrary random locations like in \cite{van2019sparse,ma2018sparse,jaritz2018sparse,wang2019plug,shivakumar2019dfusenet}, where we use nearest neighbor interpolation for the transpose sampling operation $\uparrow$  in \eqref{eq:gzupd}. The other case is \emph{depth up-sampling}, where measurements are on a regular lower-resolution grid. Given their regularity, we are able to use bi-linear interpolation for the transpose operation $\uparrow$.

\paras{Depth Un-cropping.} We next consider applications where the available measurements are dense in a contiguous (but small) portion of the image---such as from a sensor with a smaller field-of-view (FOV), or alone a single line~\cite{liao2017sparse}. In this case, we define $\mathbf{F}$ and $\mathbf{W}$ are set to measured values and one at measured locations, and zero elsewhere. We use these to define a per-patch cost $C_i(\cdot)$ for use in \eqref{eq:genopt} as
\begin{equation}
  \label{eq:uccost}
  C_i(\patch{}_i) = \lambda \|\crop{}\mathbf{W}\circ(\crop{}\Z - \crop{}\mathbf{F})\|^2,
\end{equation}
where the weight $\lambda$ is determined on a validation set.

\subsubsection{Incorporating User Guidance}

Depth estimates are often useful in interactive image editing and graphics applications. We consider a couple of settings where our estimation method can be used to include feedback from a user in the loop for improved depth accuracy.

\paras{Diverse Estimates for User Selection.} We use Batra \etal's approach~\cite{batra2012diverse} to derive multiple diverse global estimates $\{\Z^1,\ldots \Z^M\}$ of the depth map $\Z$ from our distribution $p(\Z|\I)$, and propose presenting these as alternatives to the user. We set the first estimate $\Z^1$ to our mean estimate, generate every subsequent estimate $\Z^{m+1}$ by finding a mode using \eqref{eq:genopt} with per-patch costs $C_i(\cdot)$ defined as
\begin{equation}
  \label{eq:mbest}
  C_i(\patch{}_i) = -\lambda/m \sum_{m'=1}^{m} \|\crop{}\Z^{m'} - \patch{}_i\|^2.
\end{equation}
This introduces a preference for samples that are different from corresponding patches in previous estimates, weighted by a scalar hyper-paramter $\lambda$ (set on a validation set).

\paras{Using Annotations of Erroneous Regions.} As a simple extension, we consider also getting annotations of regions with high error from the user, in each estimate $\Z^m$. Note that we only get the locations of these regions, not their correct depth values. Given this annotation, we define a mask $\mathbf{W}^M$ that is one within the region and zero elsewhere, and now recover each $\Z^{m+1}$, with a modified cost $C_i(\cdot)$:
\begin{equation}
  \label{eq:obmbest}
  C_i(\patch{}_i) = -\lambda/m \sum_{m'=1}^{m} \|(\crop{}\mathbf{W}^{m'})\circ(\crop{}\Z^{m'} - \patch{}_i)\|^2,
\end{equation}
where $\circ$ denotes element-wise multiplication, and the masks focuses the cost on regions marked as erroneous.

%%%%%%%%%%%%%%%%%%%%%%%%%%%%%%%%%%%%%%%%%%%%%%%%%%%%%%%%%%%%%%%%%
\newcommand{\colorize}{Levin \cite{levin2004colorization}}
\newcommand{\solver}{Solver \cite{barron2016fast}}
\newcommand{\stod}{Ma \cite{ma2018sparse}}

\definecolor{appname}{rgb}{0.75,0.75,0.95}
\definecolor{tspcolor}{rgb}{0.85,0.85,0.85}

\newcommand{\tspec}{\rowcolor{tspcolor}\cellcolor{white}&}
\renewcommand{\midrule}{\specialrule{0.1pt}{0pt}{.6pt}}
\renewcommand{\cline}[1]{\cmidrule{#1}}

\renewcommand{\ourso}{\em Ours}
\begin{table*}[!t]{\small
    \begin{center}
      \parbox[t]{0.515\textwidth}{\setlength\tabcolsep{4pt}\begin{tabular}{clcccccc}
    \toprule
    \multirow{2}{*}{Setting} & \multirow{2}{*}{Method} &\multicolumn{3}{c}{lower is better} & \multicolumn{3}{c}{higher is better} \\ \cline{3-8}
    &  & rms & m-rms  & rel& $\delta_{1}$ & $\delta_{2}$ & $\delta_{3}$  \\\toprule

    \rowcolor{appname}\multicolumn{8}{l}{\bf Monocular Depth Estimation}\vspace{0.25em}\\
     & Lee~\cite{lee2019monocular} & 0.538     & 0.470     & 0.131     & 83.7     & 97.1     & \bf 99.4\\ 
     & DORN~\cite{fu2018deep}      & 0.545     & 0.462     & \bf 0.114 & 85.8     & 96.2     & 98.7\\
     & \ourso                      & \bf 0.512 & \bf 0.433 &     0.116 & \bf 86.1 & \bf 96.9 & 99.1
     \vspace{0.25em}\\

%%%%%%%%%%%%%

    \rowcolor{appname}\multicolumn{8}{l}{{\bf Depth Un-cropping} (Setting = measurement FOV)}\vspace{0.25em}\\

    \tspec Liao \cite{liao2017sparse} &     0.442 &       -   &     0.104 &     87.8 &     96.4 &     98.9\\
     Horiz. & \colorize               &     1.003 &     0.852 &     0.281 &     63.8 &     83.2 &     92.3\\
     Line& Wang~\cite{wang2019plug}   &     0.482 &     0.394 &     0.089 &     90.7 &     97.3 &     99.1\\
     &  \ourso                        & \bf 0.431 & \bf 0.356 & \bf 0.088 & \bf 91.1 & \bf 98.1 & \bf 99.5\\

     \midrule

     $\ ^*$120$~~$ & \colorize     &     1.104 &     0.953 &     0.348 &     57.5 &     79.2 &     90.0\\
     x & Wang~\cite{wang2019plug}  &     0.493 &     0.409 &     0.097 &     89.1 &     96.9 &     98.9\\
     160& \ourso                   & \bf 0.447 & \bf 0.374 &\bf  0.097 &\bf  89.5 & \bf 97.7 &\bf  99.3\\

     \midrule

     $\ ^*$240$~~$ & \colorize     &     0.664 &     0.578 &     0.196 &     74.2 &     91.8 &     96.7\\
     x & Wang~\cite{wang2019plug}  &     0.416 &     0.342 &     0.081 &     91.5 &     97.7 &     99.2\\
     320&  \ourso                  & \bf 0.363 & \bf 0.298 & \bf 0.076 & \bf 92.5 & \bf 98.3 &\bf  99.5\\
     \multicolumn{8}{l}{\footnotesize $\ ^{*}$ Metrics computed only on filled-in regions.}\\

%%%%%%%%%%%%%

     \rowcolor{appname}\multicolumn{8}{l}{{\bf Depth Up-sampling} (Setting = $\uparrow$ factor)}\vspace{0.25em}\\

     \tspec Chen~\cite{chen2018estimating} &     0.318 &       -   &     0.061 &     94.2 &     98.9 & \bf 99.8\\
     \multirow{2}{*}{96x} & \colorize      &     0.512 &     0.443 &     0.120 &     85.9 &     97.1 &     99.4\\
     & Wang~\cite{wang2019plug}            &     0.367 &     0.296 &     0.057 &     95.4 &     98.7 &     99.6\\
     &  \ourso                             & \bf 0.313 & \bf 0.259 & \bf 0.056 & \bf 95.7 & \bf 99.2 & \bf 99.8\\
    
     \midrule
    
     \tspec Chen~\cite{chen2018estimating} & \bf 0.193 &       -   & \bf 0.032 & \bf 98.3 & \bf 99.7 & \bf 99.9\\
     \multirow{2}{*}{48x} & \colorize      &     0.319 &     0.275 &     0.065 &     95.4 &     99.1 &     99.8\\
     & Wang~\cite{wang2019plug}            &     0.318 &     0.256 &     0.048 &     96.7 &     99.2 &     99.8\\
     &  \ourso                             &     0.235 & \bf 0.195 &     0.035 &     97.7 &     99.6 & \bf 99.9\\
    
     \bottomrule

%%%%%%%%%%%%%%%%%%%%%%%%%%%%%%%%%%%%%%%%%%%%%%%%%%%%%%%%%%%%%                                                                    
  \end{tabular}}\parbox[t]{0.485\textwidth}{\setlength\tabcolsep{4pt}\begin{tabular}{clcccccc}
%%%%%%%%%%%%%%%%%%%%%%%%%%%%%%%%%%%%%%%%%%%%%%%%%%%%%%%%%%%%%                                                                    
    \toprule
    \multirow{2}{*}{Setting} & \multirow{2}{*}{Method} &\multicolumn{3}{c}{lower is better} & \multicolumn{3}{c}{higher is better} \\ \cline{3-8}
    &  & rms & m-rms  & rel& $\delta_{1}$ & $\delta_{2}$ & $\delta_{3}$  \\\toprule

    \rowcolor{appname}\multicolumn{8}{l}{{\bf Arbitrary Sparse Measurements} (Setting = \#measurements)}\vspace{0.25em}\\
    
     \tspec \stod                    &       -   &     0.351 &     0.078 &     92.8 &     98.4 &     99.6\\
     \multirow{2}{*}{20} & \colorize &     0.703 &     0.602 &     0.175 &     75.5 &     93.0 &     97.9\\
     & Wang~\cite{wang2019plug}      &     0.399 &     0.322 & \bf 0.065 & \bf 94.2 &     98.4 &     99.5\\
     & \ourso                        & \bf 0.359 & \bf 0.298 &     0.068 &     94.1 & \bf 98.8 & \bf 99.7\\
    
     \midrule
    
     \tspec \stod                    &       -   &     0.281 &     0.059 &     95.5 &     99.0 &     99.7\\
     \multirow{2}{*}{50} & \colorize &     0.507 &     0.436 &     0.117 &     86.4 &     97.1 &     99.3\\
     & Wang~\cite{wang2019plug}      &     0.364 &     0.291 & \bf 0.056 &     95.5 &     98.8 &     99.6\\
     & \ourso                        & \bf 0.320 & \bf 0.262 & \bf 0.056 & \bf 95.6 & \bf 99.1 & \bf 99.8\\
    
     \midrule
    
     \multirow{3}{*}{100}& \colorize &     0.396 &     0.340 &     0.085 &     92.2 &     98.5 &     99.6\\
     & Wang~\cite{wang2019plug}      &     0.336 &     0.271 &     0.052 &     96.2 &     99.0 &     99.7\\
     & \ourso                        & \bf 0.279 & \bf 0.231 & \bf 0.046 & \bf 96.6 & \bf 99.4 & \bf 99.9\\
    
     \midrule
    
     \tspec \stod                    &       -   &     0.230 &     0.044 &     97.1 &     99.4 &     99.8\\
     \multirow{2}{*}{200}& \colorize &     0.305 &     0.264 &     0.061 &     95.7 &     99.2 &     99.8\\
     & Wang~\cite{wang2019plug}      &     0.316 &     0.254 &     0.048 &     96.6 &     99.2 &     99.6\\
     & \ourso                        & \bf 0.246 & \bf 0.203 & \bf 0.039 & \bf 97.4 & \bf 99.5 & \bf 99.9

      \vspace{0.45em}\\

    \rowcolor{appname}\multicolumn{8}{l}{{\bf User Selection} (Setting = \#choices)}\vspace{0.25em}\\
     5 &  \ourso &  0.471 & 0.406 & 0.113 & 87.1 & 97.4 & 99.3 \\ \midrule
     10 &  \ourso & 0.457 & 0.394 & 0.109 & 87.9 & 97.6 & 99.4 \\ \midrule
     15 &  \ourso & 0.447 & 0.385 & 0.108 & 88.3 & 97.8 & 99.4
     \vspace{0.75em}\\

    \rowcolor{appname}\multicolumn{8}{l}{{\bf User Selection with Annotation} (Setting = \#choices)}\vspace{0.25em}\\
     5 &  \ourso &  0.398 & 0.342 & 0.098 & 90.4 & 98.2 & 99.6 \\ \midrule
     10 &  \ourso & 0.372 & 0.322 & 0.093 & 91.5 & 98.5 & 99.7 \\ \midrule
     15 &  \ourso & 0.364 & 0.315 & 0.090 & 91.9 & 98.7 & 99.7 \\\bottomrule

  \end{tabular}}
\end{center}
}
  \caption{Results for various applications on the NYUv2 test set. We use distributional outputs from our common model to generate depth estimates in a diverse variety of application settings: from standard monocular estimation to several applications when different forms of additional depth cues are available. We compare to other methods for these applications, including those (shaded background) dependent on task-specific networks trained separately for each setting. Our network, in contrast, is task-agnostic and trained only once.}
  \label{tab:bigtable}
\end{table*}

%%%%%%%%%%%%%%%%%%%%%%%%%%%%%%%%%%%%%%%%%%%%%%%%%%%%%%%%%%%%%%%%%

\subsection{Other Inference Tasks}

Our distributional output is versatile and can be used to perform general inference tasks, not just estimate per-pixel depth. We describe two such applications below.

\paras{Confidence-guided Sampling.} We can use $p(\Z|\I)$ to compute a per-pixel variance map,  as the variance of each pixel's depth value across patches and samples in $\{\Phi_i\}$ (which differs from the actual variance under $p(\Z|\I)$ by a constant $h^2$). This gives us spatial map of the relative monocular ambiguity in depth at different locations. When seeking to estimate depth from arbitrary sparse measurements, we can use this map to select where to make measurements (assuming the depth sensor provides such control). Specifically, given a budget on the total number of measurements, we propose choosing an optimal set of measurement points as local maxima of the variance map.

\paras{Pair-wise Depth.} A useful monocular depth inference task, introduced in \cite{zoran2015learning}, is to predict the ordinal relative depth of pairs of nearby points in the scene: whether the points are at similar depths (within some threshold), and if not, which point is nearer. We use our distributional output to solve this task, by looking at the relative depth in all samples in all patches that contain a pair of queried points, outputting the ordinal relation that is most frequent. We find this leads to more accurate ordinal estimates, in comparison to simply using the ordering of the individual depth value pairs in a monocular depth map estimate (as done in \cite{chen2016single,zoran2015learning}). 

\section{Experiments}
\label{sec:apps}

We now evaluate our approach on the NYUv2 dataset~\cite{silberman2012indoor} by training a common task-agnostic distributional monocular model and applying it to solve a diverse range of inference tasks in various application settings.

\paras{Preliminaries.} We use raw frames from scenes in the official train split for NYUv2~\cite{silberman2012indoor} to construct train and val sets, and report performance on the official test set. We use feature extraction layers from a pre-trained DORN model~\cite{fu2018deep}, and since it operates on inputs and outputs rescaled to a lower resolution (to $257\times 353$ from $640\times 480$), we do the same for our VAE. However, our outputs are rescaled back to the orginal full resolution to compute errors. Input depth measurements, if any, are also provided at full resolution (see supplementary). We use overlapping patches of size $33\times 33$ with stride four, and generate 100 samples per-patch to construct $\{\Phi_i\}$. Generating samples takes 5.7s on a 1080Ti GPU for each image, while inference from these samples is faster (see supplementary). Our code and trained model will be made available on publication.

\begin{figure*}[t]
  \centering
  \includegraphics[width=\textwidth]{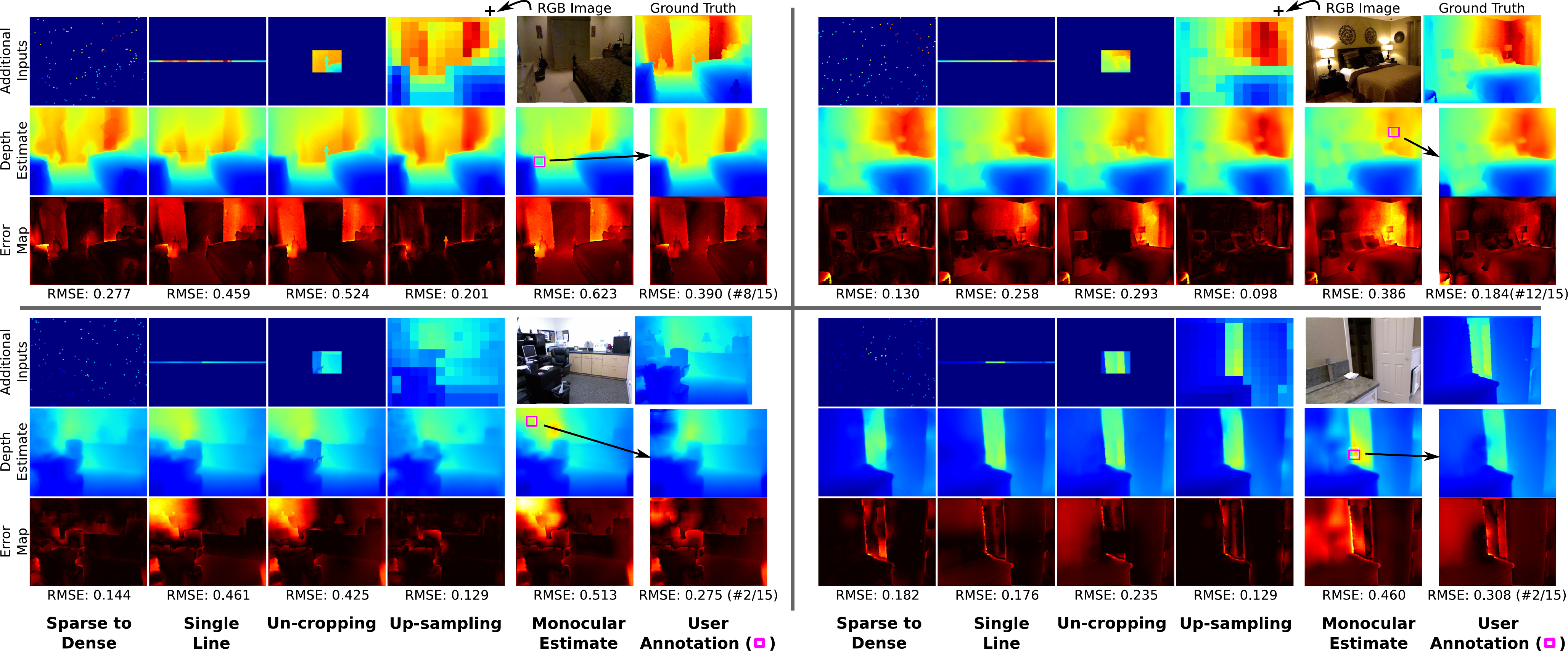}
  \caption{Example depth estimates for different applications. We show outputs from our method for both the pure monocular setting, as well as the improved estimates we obtain combining our distributional output with additional depth information---such as different kinds of partial measurements, and user guidance with annotation and selection.}
  \label{fig:results}
\end{figure*}

\subsection{Performance on Various Inference Tasks}

We evaluate depth estimation using our common model for several applications, and report performance in terms of standard error metrics on the official NYUv2 test set (see \cite{eigen2015predicting})\footnote{Some papers interpret RMSE as mean of per-image RMSE values. We report the standard definition as rms, and this per-image version as m-rms.} in Table~\ref{tab:bigtable}. We report performances on standard monocular estimation, as well for the different depth completion and user guided applications described in Sec.~\ref{sec:addinf}. We simulate user-guidance using ground-truth depth---selection of a global depth map is done automatically based on lowest error, and annotation by choosing $50\times 50$ windows with the highest error against the ground truth and no more than 50\% overlap with previously marked regions.

Not only does our method perform well in the monocular setting---outperforming the DORN~\cite{fu2018deep} whose features it uses---it is able to improve upon this monocular estimate with different available depth cues in the various applications. We find sparse measurements are most complementary to the monocular cue, and that user annotation is more useful than selection alone. Figure~\ref{fig:results} shows example depth reconstructions by our method for several applications.

Table~\ref{tab:bigtable} provides comparisons to a number of other depth completion methods. Two of these do not require task-specific training---Levin \etal's colorization method~\cite{levin2004colorization}, and Wang \etal's~\cite{wang2019plug} approach to back-propagating errors from measurements. As Wang \etal's own results were with older monocular networks, for a fairer comparison, we derive improved results by applying their method on the same DORN~\cite{fu2018deep} model as used by our network (finding optimal settings on a val set). As seen in Table~\ref{tab:bigtable}, our approach is more accurate than both these methods.

We also compare to application-specific approaches that train specialized networks separately for each application (and each setting). For depth completion from sparse measurements, we compare to the work of Chen \etal~\cite{chen2018estimating} for measurements on a regular grid, and of Ma \etal~\cite{ma2018sparse}\footnote{\cite{ma2018sparse} uses a non-standard resolution and crop to evaluate their method and report errors. We report our performance with official settings here be consistent with the benchmark and the other applications. Our performance under \cite{ma2018sparse}'s settings is similar, and reported in the supplementary.}  for those at random locations. For estimation from horizontal line measurements, we show comparisons to the method by Liao \etal~\cite{liao2017sparse}\footnote{\cite{liao2017sparse} uses measurements along a line simulated to be horizontal in 3D, leading to different $y$ image co-ordinates for each $x$. Lacking exact details for replicating their setting, we use the same number of measurements but from a line that is horizontal simply in the image plane.}. We find that our results---from a common task-agnostic network model---are comparable, and indeed often better, than these application-specific methods.

\begin{table}[!t]
  \begin{center}\small
  \setlength\tabcolsep{10pt}
  \begin{tabular}{lcccc}
    \toprule
    Measurements & 20 & 50 & 100 & 200 \\ \midrule
    Random & 0.359 & 0.320 & 0.279 & 0.246 \\
    \em Guided & \bf 0.331 & \bf 0.286 & \bf 0.253 & \bf 0.227 \\\bottomrule
             
  \end{tabular}
  \end{center}
  \caption{RMS error for depth estimation from different numbers of sparse measurements, when making measurements at random locations vs.\ with guidance from our distribution. Given the measurements, we use our depth completion approach in both cases.}
  \label{tab:opt}
\end{table}
\begin{table}[!t]
  \begin{center}\small
  \setlength\tabcolsep{4pt}
  \begin{tabular}{lccc}
    \toprule
    Method & WKDR~~ & ~WKDR$^{=}$ & ~WKDR$^{\neq}$ \\ \toprule
    Zoran \cite{zoran2015learning} & 43.5\% & 44.2\% & 41.4\% \\
    Chen \cite{chen2016single} & 28.3\% & 30.6\% & 28.6\% \\
    Xian \cite{xian2018monocular} & 29.1\% & 29.5\% & 29.7\% \\
    Ours: mean  & 30.2\% & 29.9\% & 30.5\% \\
    \em Ours (distribution) & \bf 27.1\% & \bf 26.0\% & \bf 27.8\% \\ \bottomrule
  \end{tabular}
  \end{center}
  \caption{Error rates for pairwise ordinal depth ordering from our common model, compared to other methods that used accurate ordering as an objective during training. We also report baseline errors from predictions just based on our mean depth estimate.}
  \label{tab:pa}
\end{table}

Next, we evaluate the efficacy of our approach to enabling applications beyond those that estimate depth maps. In Table~\ref{tab:opt}, we report results for making sparse depth measurements guided by the color image using our approach for different budgets on the number of measurements. Our guided measurements lead to better dense depth estimates than those at random locations (given measurements, we use our depth estimation algorithm in both cases).

Finally, we evaluate using our distribution to predict pairwise depth ordering in Table~\ref{tab:pa}, comparing it to three methods that specifically target this task: \cite{chen2016single,zoran2015learning,xian2018monocular}. Results are reported in terms of the WKDR error metrics, on a standard set of point pairs on the NYUv2 test set (see \cite{zoran2015learning}). We find that using our method leads to better predictions than from these methods, and that using our distributional output is crucial---since the accuracy of simply using the orderings from our monocular mean estimate is much lower.

\subsection{Analysis and Ablation}

\begin{figure}[!t]
  \centering
  \includegraphics[width=\columnwidth]{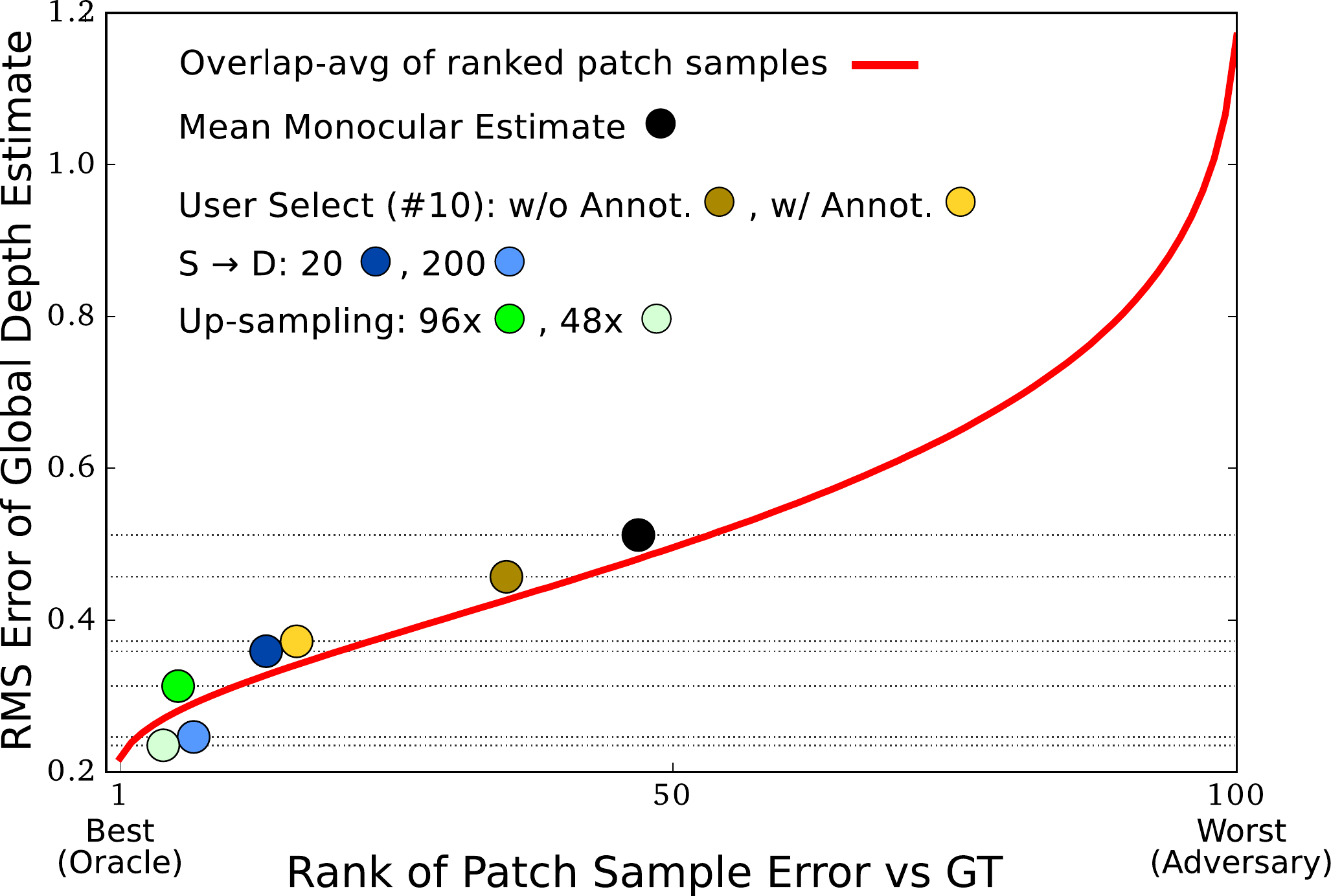}
  \caption{Analysis of distributional output and inference method on the test set. Our distribution allows for many possible global depth explanations, visualized here by choosing one of the generated samples in each patch based on the rank of its accuracy going from best (oracle) to worst (adversary), and computing global depth by overlap-average. These solutions span a large range in accuracy, and without any additional information, the mean monocular estimate lies in the middle of this range. But when additional cues are available, they can be effectively exploited by our \emph{MAP} estimation method to extract better solutions from our distribution.}
  \label{fig:squal}
\end{figure}

We visualize the diversity of depth hypotheses in our distribution in Fig.~\ref{fig:squal}. We choose one sample for each patch---based on its rank among samples for that patch in terms of accuracy relative to ground-truth. We vary this rank from best to worse, form a global depth map for each rank by overlap-average, and plot the resulting accuracies. Given the ambiguity of the monocular cue, these span a diverse range---from a very accurate estimate when an oracle allows ideal selection, to higher errors when adversarially choosing the worst samples in every patch.

Figure~\ref{fig:squal} also overlays the performance of several our inference tasks from Table~\ref{tab:bigtable}. As expected, the accuracy of pure monocular estimation is roughly at the center of the distirbution range. But when additional depth cues are available, we see that our results begin to shift to have higher accuracy---by different amounts for different applications. This shows that our inference method is successful in incorporating the information present in these depth cues.

We also study different variations to our approach for generating samples for our distribution $p(\Z|\I)$ in Table~\ref{tab:psz}---measuring  performance, on a validation set, in terms of accuracy for a ground truth-based oracle as described above, and more realistically, accuracy at monocular estimation and depth completion (from 100 measurements).

First, we evaluate using a conditional GAN~\cite{mirza2014conditional} instead of a VAE (see supplementary for architecture details). While the VAE performs better, results with the GAN are also reasonable---suggesting that our approach is compatible with different network-based sampling approaches.

Then, we consider varying the size of our patches (and proportionally, the stride). We find smaller patches actually helps oracle performance, since with the same number of samples, it is easier to generate a sample close to the ground-truth in a lower-dimensional space. However, smaller patches do not accurately capture the spatial dependencies within a patch, leading to poorer performance for actual inference. Conversely, while a higher patch size could allow encoding longer range spatial dependencies, doing so is harder via approximation from a reasonable number of samples---leading to lower accuracy both with the oracle and during inference.

For our chosen patch-size, we also evaluate higher strides, and thus lower overlap. This leads to lower performance (on depth completion), highlighting the utility of patch-overlap in the global distribution $p(\Z|\I)$, and in propagating information during inference.

\begin{table}[!t]
  \begin{center}\small
    \parbox[t]{0.72\columnwidth}{
    \setlength\tabcolsep{3pt}\small
  \begin{tabular}{llcccc}
    \toprule
    & & Oracle & Mean & S$\rightarrow$D \\ \toprule
    C-GAN& p=33,s=4 & 0.384 & 0.597 & 0.428 \\ \midrule
    C-VAE& p=17,s=2 & \bf 0.263 & 0.518 & 0.413 \\
    \bf C-VAE& \bf p=33,s=4 & 0.323 & \bf 0.516 & \bf 0.377 \\
    C-VAE& p=65,s=8 & 0.474 & 0.522 & 0.389 \\ \bottomrule
  \end{tabular}}\parbox[t]{0.28\columnwidth}{
    \setlength\tabcolsep{3pt}\small
  \begin{tabular}{cc}
    \toprule
    C-VAE &\multirow{2}{*}{S$\rightarrow$D}\\
    p=33 & \\\toprule
    s=8 & 0.396 \\
    s=16 & 0.405 \\
    s=32 & 0.436 \\\bottomrule
  \end{tabular}}
  \end{center}
  \caption{Ablation study on validation set. We evaluate different ways of generating samples: using a GAN instead of a VAE, and using different patch-sizes p (with proportional strides s). For each case, we compare achievable accuracy of individual samples via the ``oracle'' estimate (see Fig.~\ref{fig:squal}), vs.\ their utility for actual inference---in the pure monocular case and with random sparse measurements (\#100). We also evaluate the importance of patch overlap by considering larger strides for our chosen model.}
  \label{tab:psz}
\end{table}

\section{Conclusion}
\label{sec:conc}

With distributional monocular outputs,  our approach enables a variety of applications without the need for repeated training. While we considered tasks directly focused on scene geometry in this paper, we are interested in exploring how our distributional outputs can be used to manage ambiguity in downstream processing---such as for re-rendering or path planning---in future work. We also believe probabilistic predictions can be useful for other low- and mid-level scene properties, like motion and reflectance.

\noindent\textbf{Acknowledgments.} This work was supported by the NSF under award no. IIS-1820693.

{\small
%\bibliographystyle{ieee_fullname}
%\bibliography{refs}

}

\clearpage

\newcommand{\conv}[1]{conv #1$\times$#1}
\newcommand{\dconv}[1]{conv\_transpose #1$\times$#1}
\pagenumbering{roman}
\onecolumn

\appendix

\section{Additional Results}

\begin{figure}[!h]
  \centering
  \includegraphics[width=\textwidth]{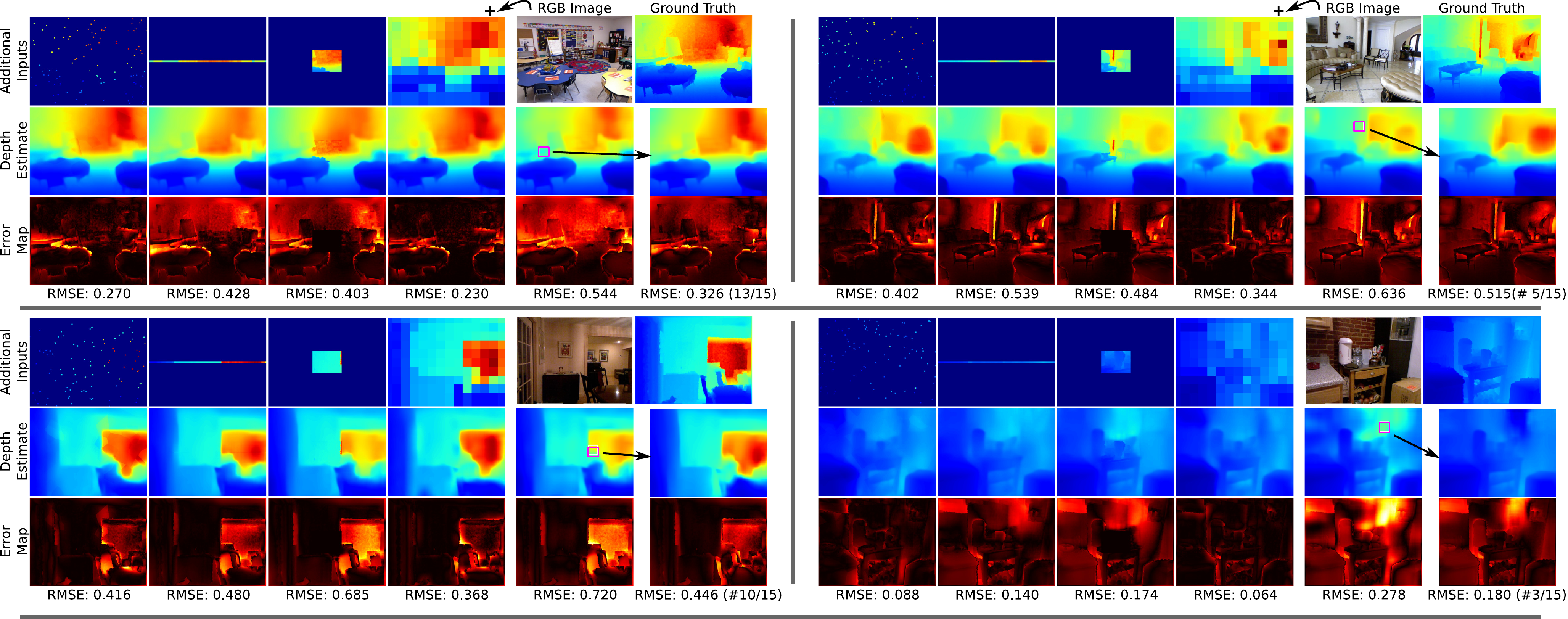}
  \includegraphics[width=\textwidth]{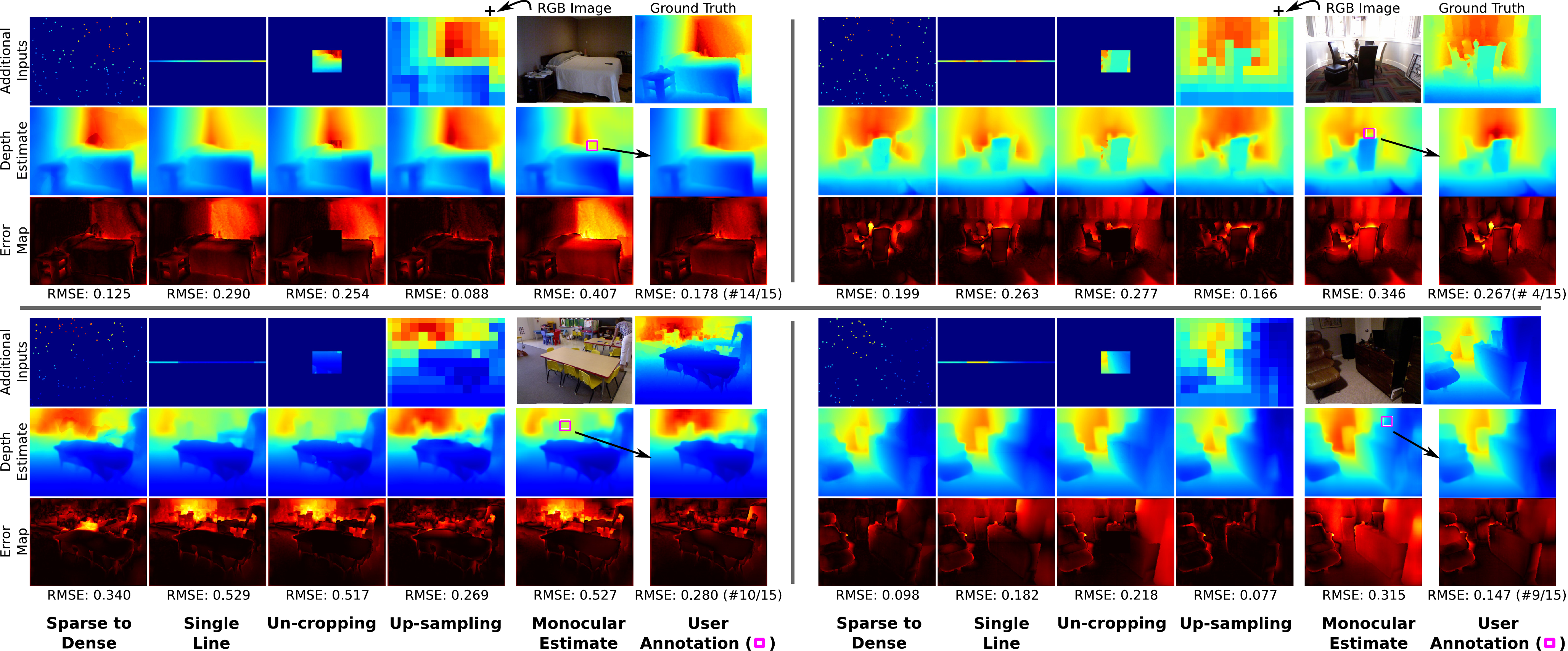} 
  \caption{Additional examples of depth reconstructions using our common model for various applications.}
  \label{fig:add}
\end{figure}

We include additional example results of depth reconstruction for various applications in Figure \ref{fig:add}.

\section{Additional Details}

\subsection{DORN Usage and Resolution}

Our conditional VAE leverages a pre-trained feature extractor derived from a state-of-the-art monocular depth network. Specifically, we take the pre-trained DORN model~\cite{fu2018deep}, remove its last two convolutional layers, and use it as our feature extractor. The DORN network works at a lower resolution of 257$\times$353, compared to the original NYUv2 resolution of 640$\times$480, for both its RGB input and depth output. Therefore, our feature extractor takes an  RGB image as input after resizing to the lower DORN resolution of 257$\times$353. The output of the feature extractor layers is a 2560-dimensional feature map with a spatial size of 33$\times$45. Our VAE takes this feature map as input, and reasons about an output depth map at the 257$\times$353 DORN resolution. We consider overlapping 33$\times$33 patches at stride 4 also at the lower DORN resolution of 257$\times$353, giving us a total of 57$\times$81 patches, each of size 33$\times$33.

Thus, our distributional output corresponds to the lower DORN~\cite{fu2018deep} resolution of 257$\times$353 for the depth map. However, all error metrics in the paper are computed (inside the valid crop) at full resolution. To do so, we resize our method's outputs to 640$\times$480 (by simple bilinear interpolation). Moreover, in all applications with additional inputs, these are also provided at the original higher resolution. For user annotations, erroneous regions are marked as 50$\times$50 windows at the full resolution, and we map the locations of these windows to the lower resolution to construct our masks $\mathbf{W}^M$. Similarly, for depth from sparse measurements, $\mathbf{F}$ corresponds to sparse measurements of depth at the full-resolution, and our global cost $C^G(\cdot)$ is defined in terms of a full-resolution depth map (we scale our depth map to the full resolution, apply gradients, and scale the updated depth map back). For depth un-cropping, we again provide depth measurements at the full resolution, and scale these to the DORN resolution to construct our measurement and mask vectors $\mathbf{F}$ and $\mathbf{W}$. Thus, all inputs and all evaluation metrics are based on the standard benchmark resolution.

\subsection{Conditional VAE Architecture}

Our conditional VAE treats the output of the DORN feature extractor---with a spatial resolution of 33$\times$45 and 2560 feature channels---as an encoding of the input image. Closely following the formulation of \cite{condVAE}, this VAE has the following three sub-networks:
\begin{enumerate}
\item \emph{Prior-net}: Given the input image feature encoding, this network produces the mean and variance vectors for each of the 57$\times$81 overlapping patches. These vectors represent the parameters of diagonal Gaussian distributions over the latent space of the corresponding patches. The latent space, and the per-patch mean and variance vectors, are 128-dimensional.
\item \emph{Encoder-decoder}: This network takes as input both the image feature encoding, and per-patch latent vectors sampled as-per the distributions produced by the prior-net.  The encoder produces a 256-dimensional feature vector for each patch (i.e., at a spatial resolution of 57$\times$81), which is then concatenated with the patch's corresponding sampled latent vector. This concatenated vector is then decoded to output 33$\times$33 depth value estimates for each patch---i.e., the output of the decoder is 57$\times$81$\times[$33$\times$33$]$. Note that the decoder path is independent for each patch to ensure independent sampling.
\item \emph{Posterior-net}: This network is used only during training, and takes the image feature encoding and ground-truth patch depths as input. It uses two streams to first encode each of these to 256-dimensional per-patch feature vectors, concatenates them, and uses a series of 1$\times$1 convolution layers to predict mean and variance vectors---the ``posterior'' equivalents of the prior-net's outputs.
\end{enumerate}

The detailed architectures of these three networks are included in Table~\ref{tab:vaearch}---with convolution and reshape operations allowing us to run the network efficiently in a fully-convolutional way, while still producing independent samples for overlapping patches. We train these three networks in a similar way as \cite{condVAE}, using a weighted combination of two losses: (1)~an $L_1$ loss between ground-truth patch depth and the output of the encoder-decoder network; and (2)~a KL-divergence loss between the distributions produced by the prior-net and posterior-net; with a weight of $1e-4$ for the latter. After training, we discard the posterior-net. Given an image, we run the prior-net and the encoder-half of the encoder-decoder network, and then run the decoder-half multiple (100) times with different samples from the latent distributions to produce multiple samples of depth estimates for each patch.

\begin{table}[!t]
  \begin{center}
    \parbox[c]{0.47\textwidth}{\centering{\scriptsize
      \begin{tabular}{|c|c|c|}
        \hline
        \em No. & \em Layer & \em Output Shape \\ \hline
        \row{0, features from feature extractor, \osize{1,33,45,2560}} \hline
        \row{1, bilinear upsample, \osize{1,65,89,2560}} \hline
        \row{2, \conv{1}, \osize{1,65,89,1024}} \hline
        \row{3, \conv{1}, \osize{1,65,89,512}} \hline
        \row{4, \conv{3} dilation=2, \osize{1,61,85,512}} \hline
        \row{5, \conv{3} dilation=2, \osize{1,57,81,256}} \hline
        \row{6, \conv{1}, \osize{1,57,81,256}} \hline
        \row{7, \conv{1}, \osize{1,57,81,256}} \hline
        \row{8, \conv{1} (no ReLU), \osize{1,57,81,256}} \hline
        \multirow{2}{*}{9} & \multirow{2}{*}{reshape and split} & \osize{(57*81),1,1,128} \ \ Mean\\
         & & \osize{(57*81),1,1,128}\ \ log-Sigma\\\hline
      \end{tabular}\\~\\}
    {\small \textbf{Prior-net}}\\~\\
    {\scriptsize
      \begin{tabular}{|c|c|c|}
        \hline
        \em No. & \em Layer & \em Output Shape \\ \hline
        \row{0a, features from feature extractor, \osize{1,33,45,2560}} \hline
        \row{1a, bilinear upsample, \osize{1,65,89,2560}} \hline
        \row{2a, \conv{1}, \osize{1,65,89,1024}} \hline
        \row{3a, \conv{1}, \osize{1,65,89,512}} \hline
        \row{4a, \conv{3} dilation=2, \osize{1,61,85,512}} \hline
        \row{5a, \conv{3} dilation=2, \osize{1,57,81,256}} \hline
        \row{6a, reshape, \osize{(57*81),1,1,256}} \hline\hline

        \rowcolor{lightgray}\row{0b, sample from latent distribution, \osize{(57*81),1,1,128}} \hline\hline
        
        \row{0, concat: 6a and 0b, \osize{(57*81),1,1,384}} \hline
        \row{1, \dconv{3}, \osize{(57*81), 3,3,256}} \hline
        \row{2, \dconv{3}, \osize{(57*81), 5,5,128}} \hline
        \row{3, \dconv{3}, \osize{(57*81), 7,7,64}} \hline
        \row{4, bilinear upsample, \osize{(57*81), 13,13,64}} \hline
        \row{5, \dconv{3}, \osize{(57*81), 15,15,32}} \hline
        \row{6, \dconv{3}, \osize{(57*81), 17,17,16}} \hline
        \row{7, bilinear upsample, \osize{(57*81), 33,33,16}} \hline

        \row{8, \conv{1} + tanh, \osize{(57*81), 33,33,1}} \hline
        \row{9, reshape, $57 \times 81 \times [33 \times 33]$} \hline
      \end{tabular}\\~\\}
    {\small \textbf{Encoder-decoder}}\\
  }
  \parbox[c]{0.45\textwidth}{\centering{\scriptsize
        \begin{tabular}{|c|c|c|}
          \hline
          \em No. & \em Layer & \em Output Shape \\ \hline
          \row{0.a, features from feature extractor, \osize{1,33,45,2560}} \hline
          \row{1.a, bilinear upsample, \osize{1,65,89,2560}} \hline
          \row{2.a, \conv{3} dilation=2, \osize{1,61,85,1024}} \hline
          \row{3.a, \conv{3} dilation=2, \osize{1,57,81,256}} \hline\hline

          \row{0.b, GT depth patches,$57\times 81\times [33\times 33]$} \hline
          \row{1.b, reshape,\osize{(57*81),33,33,1}} \hline
          \row{2.b, \conv{3} stride=2, \osize{(57*81),16,16,8}} \hline
          \row{3.b, \conv{2} stride=2, \osize{(57*81),8,8,16}} \hline
          \row{4.b, \conv{2} stride=2, \osize{(57*81),4,4,32}} \hline
          \row{5.b, \conv{2} stride=2, \osize{(57*81),2,2,64}} \hline
          \row{6.b, reshape, \osize{(57*81),1,1,256}} \hline
          \row{7.b, reshape, \osize{1,57,81,256}} \hline\hline

          \row{0 , concat: 3.a and 7.b, \osize{1,57,81,512}} \hline
          \row{1, \conv{1}, \osize{1,57,81,1024}} \hline
          \row{2, \conv{1}, \osize{1,57,81,512}} \hline
          \row{3, \conv{1}, \osize{1,57,81,256}} \hline
          \row{4, \conv{1} (no ReLU), \osize{1,57,81,256}} \hline
          \multirow{2}{*}{5} & \multirow{2}{*}{reshape and split} & \osize{(57*81),1,1,128} \ \ Mean\\
              & & \osize{(57*81),1,1,128}\ \ log-Sigma\\\hline
        \end{tabular}\\~\\}
      {\small \textbf{Posterior-net}\\}
    }
    \end{center}
  \caption{Conditional VAE architecture. We show architecture details of the three different sub-networks in our VAE, with the posterior-net used only during training. Valid padding is used everywhere. Every convolutional layer is followed by a ReLU, unless otherwise specified. The output of the encoder-decoder network has a tanh activation, followed by scaling to map to the depth range of the NYUv2 dataset.}
  \label{tab:vaearch}
\end{table}

\subsection{Inference Hyperparameter Selection}
For applications with a per-patch cost $C_i(\cdot)$---i.e., user-guidance and depth un-cropping---the value of $\lambda$ is chosen based on a small validation set, as $\lambda=10$ for user-guidance, and $150$ for un-cropping. Moreover, for user guidance, we find that slowly increasing the value of $\lambda$ from $5$ to its final value of $10$ during optimization leads to convergence to better solutions. For depth completion from sparse (both random and regularly spaced) measurements, we set the value of the parameters for gradient-based updates for the global cost---step-size $\gamma$ (in range $[0.1,1.0]$) and number of steps (in range $[1,10]$)---based on a validation set as well.

\subsection{Running Time}
Our method works by first generating multiple (100) samples for each overlapping patch, and then carrying out inference using these samples for different applications. In particular, for ``MAP'' estimation to compute depth estimates with additional information, this involves running our iterative optimization method. We report these running time (on a 1080Ti GPU) for this optimization for different applications in Table~\ref{tab:runtime}---these times vary both because of variance in time taken per-iteration, and number of iterations needed for convergence.

\begin{table}[!t]\small
  \begin{center}
     \centering
     \begin{tabular}{lccccc}\toprule
       \multirow{2}{*}{Application}&\multirow{2}{*}{Un-cropping}&\multirow{2}{*}{Up-sampling}&Sparse&\multirow{2}{*}{User Sel.}&User Sel.\\
                                   & & & Meaus.& &w/ Annot.\\\midrule
       Time & 1.0 s & 0.4 s & 0.7 s & 0.8 s & 2.2 s\\\bottomrule
     \end{tabular}
   \end{center}
     \caption{Optimization running time for different applications (does not include sample generation time). Note that for user-guidance, the reported time is for each generated mode $\mathbf{Z}^m$.}
  \label{tab:runtime}
\end{table}

\subsection{Ablation: Conditional GAN architecture}

Our conditional GAN architecture features generator and discriminator networks, with similar architecture design choices to the VAE---the generator has a similar architecture the encoder-decoder network in the VAE, and the discriminator to the posterior-net. The generator uses dropout as the noise-source to enable sampling in different runs of the generator---and ensure that per-patch estimates are independent by ensuring that different patches are based on different instantiations of dropout noise values. The architecture is detailed in Table~\ref{tab:gen}.
\begin{table}[!t]
  \begin{center}
    \parbox[c]{0.47\textwidth}{\centering{\scriptsize
        \begin{tabular}{|c|c|c|}
          \hline
          \em No. & \em Layer & \em Output Shape \\ \hline
          \row{0, features from feature extractor, \osize{1,33,45,2560}} \hline
          \row{1, resize, \osize{1,65,89,2560}} \hline
          \row{2, \conv{1}, \osize{1,65,89,1024}} \hline
          \row{3, \conv{1}, \osize{1,65,89,512}} \hline
          \row{4, \conv{3} dilation=2, \osize{1,61,85,512}} \hline
          \row{5, \conv{3} dilation=2, \osize{1,57,81,256}} \hline
          \row{6, reshape, \osize{(57*81),1,1,256}} \hline

          \rowcolor{lightgray} & \multicolumn{2}{c|}{dropout as noise} \\ \hline
          \row{7, \conv{1}, \osize{(57*81),1,1,256}} \hline
          \rowcolor{lightgray} & \multicolumn{2}{c|}{dropout as noise} \\ \hline
          \row{8, \conv{1}, \osize{(57*81),1,1,256}} \hline
          \rowcolor{lightgray} & \multicolumn{2}{c|}{dropout as noise} \\ \hline
          \row{9, \conv{1}, \osize{(57*81),1,1,256}} \hline
          \rowcolor{lightgray} & \multicolumn{2}{c|}{dropout as noise} \\ \hline

          \row{10, \dconv{3}, \osize{(57*81), 3,3,256}} \hline
          \row{11, \dconv{3}, \osize{(57*81), 5,5,128}} \hline
          \row{12, \dconv{3}, \osize{(57*81), 7,7,64}} \hline
          \row{13, resize, \osize{(57*81), 13,13,64}} \hline
          \row{14, \dconv{3}, \osize{(57*81), 15,15,32}} \hline
          \row{15, \dconv{3}, \osize{(57*81), 17,17,16}} \hline
          \row{16, resize, \osize{(57*81), 33,33,16}} \hline\hline

          \row{17, \conv{1} + tanh, \osize{(57*81), 33,33,1}} \hline
          \row{18, reshape, $57 \times 81 \times [33 \times 33]$} \hline
        \end{tabular}\\~\\}
      {\small \textbf{Generator}}}
    \parbox[c]{0.47\textwidth}{\centering{\scriptsize
        \begin{tabular}{|c|c|c|}
          \hline
          \em No. & \em Layer & \em Output Shape \\ \hline
          \row{0.a, features from feature extractor, \osize{1,33,45,2560}} \hline
          \row{1.a, resize, \osize{1,65,89,2560}} \hline
          \row{2.a, \conv{3} dilation=2, \osize{1,61,85,1024}} \hline
          \row{3.a, \conv{3} dilation=2, \osize{1,57,81,256}} \hline
          \row{4.a, reshape, \osize{(57*81),1,1,256}} \hline\hline

          \row{0.b, true/fake depth patches,$57\times 81\times [33\times 33]$} \hline
          \row{1.b, reshape,\osize{(57*81),33,33,1}} \hline
          \row{2.b, \conv{3} stride=2, \osize{(57*81),16,16,8}} \hline
          \row{3.b, \conv{2} stride=2, \osize{(57*81),8,8,16}} \hline
          \row{4.b, \conv{2} stride=2, \osize{(57*81),4,4,32}} \hline
          \row{5.b, \conv{2} stride=2, \osize{(57*81),2,2,64}} \hline
          \row{6.b, reshape, \osize{(57*81),1,1,256}} \hline\hline

          \row{0 , concat: 4.a and 6.b, \osize{(57*81),1,1,512}} \hline
          \row{1, \conv{1}, \osize{(57*81),1,1,1024}} \hline
          \row{2, \conv{1}, \osize{(57*81),1,1,512}} \hline
          \row{3, \conv{1}, \osize{(57*81),1,1,256}} \hline
          \row{4, \conv{1} + sigmoid, \osize{(57*81),1,1,1}} \hline
        \end{tabular}\\~\\}
    {\small \textbf{Discriminator}}}
  \end{center}
  \caption{Conditional GAN architecture. We show architectures for the generator and discriminator for the GAN used in our ablation study, which follows a similar overall design as our VAE. For all dropout layers, we use probability $0.5$. Every convolutional layer is followed by a ReLU, unless otherwise specified, and valid padding is used everywhere.}
  \label{tab:gen}
\end{table}

\subsection{Half-resolution Comparison to Ma \etal~\cite{ma2018sparse}.} Note that \cite{ma2018sparse} evaluate their methods by reporting errors on a centered crop of half-resolution depth-maps, and also derive their input sparse measurements at this half-resolution. In contrast, our results in Table 1 in the paper represent the official benchmark metrics (in the valid crop at full resolution) for consistency to other evaluations---in our paper and elsewhere. For a more direct comparison to \cite{ma2018sparse}, we also evaluated our method by replicating their setting. Specifically, to provide input sparse measurements, we first down-sample the ground-truth depth map and randomly sample depth values from this down-sampled map. We then provide these as inputs to our method (which resizes these back to the full resolution to compute the global cost $C^G(\cdot)$). Then, we take the full-resolution depth map estimates produced by our method, down-sample them to half-resolution, and compute error metrics on the same centered crop as \cite{ma2018sparse}. We report these results in Table~\ref{tab:s2dhalf}, and find they are similar to standard evaluation in Table 1 in the paper.

\begin{table}[!t]
  \begin{center}{\small
      \begin{tabular}{clcccccc}
        \toprule
        \multirow{2}{*}{Setting} & \multirow{2}{*}{Method} &\multicolumn{3}{c}{lower is better} & \multicolumn{3}{c}{higher is better} \\ \cline{3-8}
                                 &  & rms & m-rms  & rel& $\delta_{1}$ & $\delta_{2}$ & $\delta_{3}$  \\\toprule
        \tspec{20} \stod                   &       -   &     0.351 &     0.078 &     92.8 &     98.4 &     99.6\\
           & \ourso                        & \bf 0.363 & \bf 0.303 & \bf 0.070 & \bf 94.0 & \bf 98.7 & \bf 99.7\\\midrule
        \tspec{50} \stod                   &       -   &     0.281 &     0.059 &     95.5 &     99.0 &     99.7\\
           & \ourso                        & \bf 0.309 & \bf 0.257 & \bf 0.053 & \bf 95.8 & \bf 99.2 & \bf 99.8\\\midrule
        \tspec{200} \stod                  &       -   &     0.230 &     0.044 &     97.1 &     99.4 &     99.8\\
           & \ourso                        & \bf 0.237 & \bf 0.196 & \bf 0.037 & \bf 97.6 & \bf 99.6 & \bf 99.9\\
        \bottomrule
\end{tabular}

  }\end{center}
  \caption{Performance on depth estimation from arbitrary sparse measurements, using the same evaluation setting as \cite{ma2018sparse} (half-resolution, evaluated on a center-crop).}
  \label{tab:s2dhalf}
\end{table}

\end{document}